\documentclass{article}

\usepackage{microtype}
\usepackage{graphicx}
\usepackage{subfigure}
\usepackage{booktabs} 

\usepackage{hyperref}

\usepackage[accepted]{sysml2019}

\sysmltitlerunning{Ternary Hybrid Neural-Tree Networks for Highly Constrained IoT Applications}

\begin{document}

\twocolumn[
\sysmltitle{Ternary Hybrid Neural-Tree Networks for Highly Constrained IoT Applications}



\sysmlsetsymbol{equal}{*}

\begin{sysmlauthorlist}
\sysmlauthor{Dibakar Gope}{to}
\sysmlauthor{Ganesh Dasika}{to}
\sysmlauthor{Matthew Mattina}{to}
\end{sysmlauthorlist}

\sysmlaffiliation{to}{Arm ML Research Lab}

\sysmlcorrespondingauthor{Dibakar Gope}{dibakar.gope@arm.com}

\sysmlkeywords{Machine Learning, SysML}

\vskip 0.3in

\begin{abstract}

Machine learning-based applications are increasingly prevalent in IoT
devices.  The power and storage constraints of these devices make it
particularly challenging to run modern neural networks, limiting the number
of new applications that can be deployed on an IoT system. A number of
compression techniques have been proposed, each with its own trade-offs.
We propose a hybrid network which combines the strengths of current neural-
and tree-based learning techniques in conjunction with ternary
quantization, and show a detailed analysis of the associated model design
space.  Using this hybrid model we obtained a $11.1\%$ reduction in the number
of computations, a $52.2\%$ reduction in the model size, and a $30.6\%$ reduction
in the overall memory footprint over a state-of-the-art keyword-spotting neural network,
with negligible loss in accuracy.

\end{abstract}
]



\printAffiliationsAndNotice{}  

\section{Introduction}
\label{intro}


Machine learning algorithms, and neural networks (NNs) in particular, are increasingly deployed in Internet-of-Things (IoT)
devices. Popular applications include speech interfaces in smart-home
devices, predictive maintenance for commercial and industrial machines,
health-monitoring in wearables, etc. However, due to the energy, power,
storage, and compute limitations of highly-constrained IoT devices, they
are frequently limited to simplistic tasks, while more
sophisticated requests are off-loaded to a more capable device or to a server. In addition to being computationally constrained,
IoT devices frequently have very little available SRAM, tend to be ``always-on'', and are often connected to constrained power sources. Because of these constraints, reducing the computation and storage required by ML models
for IoT applications is of paramount importance in order to ensure a longer
battery life.

To enable this computation and size compression of NN models, one
particularly effective technique has been the use of depthwise-separable
(DS) convolutional layers. We see these layers being used in a large, image
classification application~\cite{mobileNets2017} and also on a ubiquitous
keyword-spotting application~\cite{helloEdge2017}, showing state-of-the-art
or near-state-of-the-art accuracy in both cases.

While DS convolutional layers have been transformative, even further
compression is still valuable in order to target the most constrained
microcontrollers or to make a wider range of applications available on IoT
devices. Recent work has shown that this might be possible through the use
of binary- and ternary-weight
networks~\cite{XNORNet2016,ternaryNN2016,ternaryWeightNetworks2016,strassennets2018}.
In such networks, multiplications are replaced with additions, relying on
binary (-1,1) or ternary (-1,0,1) weight matrices. This enables more energy-efficient and faster network architectures with fewer expensive
multiplications but at the cost of modest to significant drop in
prediction accuracy when compared to their full-precision counterparts. Recent work on
StrassenNets~\cite{strassennets2018} presents a more mathematically
profound way to approximate matrix multiplication computation (and, in
turn, convolutions) using mostly ternary weights and a few full-precision
weights. It demonstrates no loss in predictive performance when compared to
full-precision models. The effectiveness of StrassenNets is quite variable,
however, depending on the neural network architecture. We observe, for
example, that while \textit{strassenifying} is effective in reducing the
model size of DS convolutional layers, this might come with a
prohibitive increase in the number of addition operations, reducing the energy
efficiency of neural network inference.

The interest in reducing complexity has also expanded beyond neural
networks. Recent research around tree-based learning algorithms has shown
immense potential to perform classification and regression in the IoT
setting with a significantly lower computation and storage budget than their
neural counterparts, while maintaining acceptable model accuracy.
More
specifically, Bonsai decision trees~\cite{bonsaitree2017} make this possible by
learning a single, shallow and sparse tree to reduce model size but with
powerful nodes for accurate prediction.
It uses more powerful branching functions than the axis-aligned hyperplanes in standard decision trees.
This is coupled with non-linear predictions made by internal and
leaf nodes on a single, shallow decision tree learned in a low-dimensional
space. Combining these ideas allows Bonsai trees to learn complex non-linear decision boundaries
using a compact representation. 
While the results in~\cite{bonsaitree2017} show
promising results for smaller applications, our observations were that the
techniques do not scale when extended to a more complex use-case, showing poor prediction accuracy even with a large model footprint. 

We now, therefore, have two ways of compressing models, each with its
own advantages and limitations:
\begin{itemize}
\item{StrassenNets is effective at reducing the size of a neural network
	model but at a potentially significant cost of more addition
		operations.}
\item{Bonsai tree is effective at reducing the number of operations for
	simple models but cannot easily be extended to larger models.}
\end{itemize}

Motivated by these observations, we propose a hybrid network architecture
capable of giving start-of-the-art accuracy levels, while requiring a
fraction of the model parameters and considerably fewer operations per
inference. 
The hybrid architecture makes this possible by leveraging a few neural DS convolutional layers for feature extraction
and then relying on a compute-efficient, shallow Bonsai decision tree to perform the classification. It then applies StrassenNets over the overall neural-tree network to reduce its memory footprint significantly thus enabling a compact compute-efficient architecture. We apply this hybrid architecture to a
representative IoT application keyword-spotting.
The hybrid network achieves a $98.89\%$ reduction in multiplications, a $12.22\%$ reduction in additions (overall $11.1\%$ reduction in number
of operations), a $52.2\%$ reduction in model size, and a $30.6\%$ reduction in overall memory footprint over a 
state-of-the-art keyword-spotting neural model. The hybrid network accomplishes this with a very minimal loss in accuracy of $0.27\%$.
The final network is well within the constrained compute budget of typical microcontrollers.

The remainder of the paper is organized as follows.
Section~\ref{sec:modelcompression} elaborates on the incentives behind this
hybrid network architecture for microcontrollers and provides a brief overview
of the neural and tree-based learning algorithms that it attempts to hybridize
along with our observations of applying them to the keyword-spotting
application. Failing to find a good balance between accuracy and computation
costs shifts our focus towards designing a hybrid neural-tree network.
Section~\ref{sec:hybridNetwork} describes our hybrid network.
Section~\ref{sec:evaluation} presents results.
Section~\ref{sec:related_work} compares our hybrid network against prior works and and Section~\ref{sec:conclusion} concludes the paper.


\section{Model Compression Limitations for an IoT Application}
\label{sec:modelcompression}

\subsection{StrassenNets}
Given two $2\times2$ matrices, Strassen's matrix multiplication algorithm computes
their product using $7$ multiplications instead of the $8$ required with a
na\"ive implementation of matrix multiplication. 
It essentially
converts the matrix multiplication operation to a 2-layer sum-product
network (SPN) computation as shown below:

\begin{equation}
vec(C) = W_c[(W_bvec(B)) \odot (W_avec(A))]
\end{equation}

$W_a$, $W_b$ $\in$ $K^{r \times n^2}$ and $W_c$ $\in$ $K^{n^2 \times r}$ are
ternary matrices with $K$ $\in$ $\{-1,0,1\}$,
$vec(A)$ and $vec(B)$ are the vectorization of the two input square matrices
$A$, $B$ $\in$ $R^{n \times n}$; and $vec(C)$ represents the vectorized form of
the product $A \times B$. $\odot$ denotes the element-wise product. The
$(W_bvec(B))$ and $(W_avec(A))$ of the SPN compute $r$ intermediate factors
each from additions, and/or subtractions of elements of $A$ and $B$ realized by
the two associated ternary matrices $W_a$ and $W_b$ respectively. 
The two generated $r$-length intermediate factors are then element-wise multiplied to produce the $r$-length $(W_bvec(B)) \odot (W_avec(A))$.
The outmost ternary matrix
$W_c$ later combines the $r$ elements of the product $(W_bvec(B)) \odot
(W_avec(A))$ in different ways to generate the vectorized form of product
matrix $C$.  Therefore, the width of the hidden layer of the SPN $r$ decides
the number of multiplications required for the Strassen's matrix multiplication
algorithm. For example, given two $2\times2$ matrices, ternary matrices $W_a$
and $W_b$ with sizes of $7\times4$ can multiply them using $7$ multiplications
and $36$ additions. It is important to note that Strasssen's algorithm requires a
hidden layer with $7$ units here to compute the exact product matrix that a na\"ive
matrix multiplication algorithm can obtain using $8$ multiplications.

Building on top of Strassen's matrix multiplication algorithm, the StrassenNets
work~\cite{strassennets2018} instead realizes approximate matrix
multiplications in DNN layers using fewer hidden layer units compared to the
standard Strassen's algorithm to achieve the exact product matrix. StrassenNets
makes this possible by training a SPN-based DNN framework end-to-end to learn
the ternary weight matrices from the training data. The learned ternary weight
matrices can then approximate the otherwise exact matrix multiplications of the
DNN layers with significantly fewer multiplications than Strassen's
algorithm. The approximate transforms realized by the SPNs, adapted to the DNN
architecture and application data under consideration, can enable precise
control over the number of multiplications and additions required per inference,
creating an opportunity to tune DNN models to strike an optimal balance between
accuracy and computational complexity.  The success of StrassenNets in
achieving significant compression for $3 \times 3$
convolutions~\cite{strassennets2018} and increasing visibility of DS
convolutions in resource-constrained IoT networks~\cite{helloEdge2017} inspired
us to apply StrassenNets over already compute-efficient IoT networks dominated
with DS layers to reduce their computational costs and model size even further.
Further compression of DS layers will not only enable more energy-efficient IoT
networks leading to longer lasting batteries, but also will open up the
opportunities for more complex IoT use-cases to fit in the limited memory
budget of tiny microcontrollers. 

As a representative benchmark for exploring different compression algorithms,
we have chosen a keyword spotting (KWS) model from~\cite{helloEdge2017}.  The
DS convolution-based model (DS-CNN) shown
in~\cite{helloEdge2017} has state-of-the-art accuracy on the realistic Google
speech commands dataset~\cite{SpeechCommandsDataset2018}. Furthermore, when
compared to traditional CNN or other RNN approaches, the model size is smaller
and the number of operations required per inference is fewer as well.

\subsubsection{StrassenNets for KWS}
\label{subsec:ST_DS-CNN_eval}

\begin{table}[t]
   \caption{Test accuracy along with the number of multiplications, additions, operations and model size for state-of-the-art DS-CNN and strassenified DS-CNN (ST-DS-CNN) on KWS. $r$ is the hidden layer width of a strassenified convolution layer, $c_{out}$ is the number of output channels of the corresponding convolution layer.}
\label{strassenKWS}
\vskip 0.15in
\begin{center}
\begin{tiny}
\begin{sc}
\begin{tabular}{lccccccr}
\toprule
Network & Acc. & Muls, & MACs & Ops & Model\\
	&  (\%) & Adds &  &  & size\\
\midrule
DS-CNN    & 94.4 & - & 2.7M & 2.7M & 22.07KB\\
ST-DS-CNN & 93.18 & 0.05M, 2.85M & - & 2.9M & 16.23KB\\
($r = 0.5c_{out}$) &  &  &  &  &\\
ST-DS-CNN & 94.09 & 0.06M, 4.09M & - & 4.15M & 19.26KB\\
($r = 0.75c_{out}$) &  &  &  &  &\\
ST-DS-CNN & 94.03 & 0.07M, 5.32M & - & 5.39M & 22.29KB\\
($r = c_{out}$) &  &  &  &  &\\
ST-DS-CNN & 94.74 & 0.11M, 10.25M & - & 10.36M & 34.42KB\\
($r = 2c_{out}$) &  &  &  &  &\\
\bottomrule
\end{tabular}
\end{sc}
\end{tiny}
\end{center}
\vskip -0.1in
\end{table}


We observe that although strassenifying DS convolution layers reduces multiplications
significantly as expected, it increases additions considerably in order to
achieve an accuracy comparable to that of the state-of-the-art DS-CNN.
Table~\ref{strassenKWS} captures our observation with strassenifying DS
	layers of the uncompressed DS-CNN KWS model. Multiply, addition, and multiply-accumulate (MAC) operations typically incur similar execution latencies in modern microprocessors, but different models have different ratios of these operations. They are, therefore, counted individually and aggregated in the ``Ops'' column.
The strassenified
network with the $r = 0.75c_{out}$ configuration incurs a negligible loss in
accuracy of $0.31\%$ while reducing multiplications by $97.7\%$ but increasing
additions by $51.4\%$ ($2.7$M MACs of DS-CNN vs. $0.06$M multiplications and
$4.09$M additions of ST-DS-CNN with $r = 0.75c_{out}$).
	That means the strassenified
network with $r = 0.75c_{out}$ configuration actually increases the
number of total operations
to $4.15$M when
compared to $2.7$M operations in the uncompressed DS-CNN network.  As shown in
Table~\ref{strassenKWS}, a number of potential values for the hidden layer
width ($r$) were explored and a value of at least $0.75c_{out}$ was needed to
achieve a comparable accuracy to that of the full-precision DS-CNN model. Using
fewer hidden units ($r = 0.5c_{out}$) than this incurs an accuracy loss of
$1.22\%$, whereas wider strassenified hidden layers ($r =
2c_{out}$) recover the negligible accuracy loss of the $r = 0.75c_{out}$
configuration.
For sufficiently large $r$ values, the strassenified network can even out-perform the uncompressed DS-CNN model
in accuracy, albeit with a significant increase (about $280\%$ for $r = 2c_{out}$) in the number of additions than the
DS-CNN model.

\subsubsection{Compute inefficiency of StrassenNets for models with DS convolutions}

It is important to note here that although the number of additions does
increase marginally with strassenifying standard $3 \times 3$ or $5 \times 5$
convolutional layers~\cite{strassennets2018}, that trend does not hold true with strassenifying DS layers. 
This stems from the fact that $1 \times 1$ pointwise convolutions dominate the compute bandwidth of a neural network with DS layers~\cite{helloEdge2017,mobileNets2017} and strassenifying a
$1 \times 1$ pointwise convolution requires executing two equal-sized (for $r = c_{out}$) $1 \times 1$ convolution operations (with ternary weight filters) in place of the standard $1 \times 1$ convolution. This results in a significant increase in additions in comparison to the execution of the standard $1 \times 1$ convolution.
In contrast to that, a $3 \times 3$ strassenified convolution with $r = c_{out}$ instead requires executing a $3 \times 3$ convolution and a $1 \times 1$ convolution with ternary weight filters, 
causing a marginal increase in additions compared to
the execution of the standard $3 \times 3$ convolution~\cite{strassennets2018}.
This overhead of addition operations with strassenified DS convolutions increases
in proportion to the width of the strassenified hidden
layers, i.e. to the size of the ternary convolution operations, as observed in Table~\ref{strassenKWS}. As a result,
a strassenified DS convolution layer may incur enough overhead to offset the benefit of strassenifying a DS convolution layer.

While~\cite{strassennets2018} demonstrates better trade-offs when
strassenifying ResNet-18 architecture, this is not likely to continue once a larger network dominated
with DS convolutions (e.g. MobileNets~\cite{mobileNets2017}) is strassenified. \cite{strassennets2018} 
observes the ResNet-18 architecture with strassenified $3 \times 3$
convolutions to achieve comparable accuracy to that of the uncompressed
ResNet-18 on the ImageNet dataset with $r = 2c_{out}$ configuration while
requiring a modest ($29.63\%$) increase in additions. A strassenified
MobileNets with $r = 2c_{out}$ configuration for the DS layers will give
rise to about a $300\%$ increase in additions over the uncompressed
MobileNets architecture.
This increase in computational costs associated with
strassenified DS convolutions in conjunction with the high accuracy and low
latency requirements of IoT applications call for a model architecture
exploration that can leverage the compute efficiency of DS layers and model
size reduction of strassenified convolutions owing to their ternary weights
while maintaining acceptable or no increase in additions. As tree-based
learning techniques from recent work~\cite{bonsaitree2017} exhibit
accuracy on par with neural models while requiring significantly fewer
MAC operations, this motivates us to explore the model accuracy and
compute-efficiency of these tree-based techniques for representative IoT
applications.

\subsection{Bonsai Decision Trees} Piece-wise axis-aligned decision boundaries
coupled with constant predictions at just the leaf nodes restrict the
prediction accuracy of typical tree models when compared to that of their neural
counterparts. Tree ensembles are commonly used to improve the accuracy, but
they can occupy too large a memory footprint for typical microcontrollers. Recent work on tree models attempt to learn more complex
decision boundaries by moving away from learning axis-aligned hyperplanes at
internal nodes and constant predictors at the leaves. Bonsai decision
trees~\cite{bonsaitree2017} fall into this paradigm. Using more powerful
branching functions than the axis-aligned hyperplanes of standard decision trees in
conjunction with non-linear prediction scores in both internal and leaf nodes
allows Bonsai to learn a single, shallow tree that can achieve accuracy on par
with small neural-based models. For a multi-class classification problem with
$L$ targets, Bonsai learns matrices $W_{\hat{D} \times L}$ and $V_{\hat{D}
\times L}$ at both leaf and internal nodes so that each node now predicts a
non-linear prediction score $W^{\top}Zx \circ tanh(\sigma V^{\top}Zx)$. Bonsai
reduces model size by projecting each $D$-dimensional input feature vector $x$
into a low $\hat{D}$-dimensional space using a projection matrix $Z_{\hat{D}
\times D}$ in which the tree is learned.  Once an input feature is projected to
a low-dimensional space, Bonsai adds the individual node predictions along the
path traversed by the projected input to derive the overall prediction.  Owing
to a single, shallow tree with powerful nodes and branching functions learned
in a low-dimensional space, Bonsai can achieve impressive computation reduction
over a typical DNN, while preserving DNN-level accuracy for very small models.


\subsubsection{Bonsai tree for KWS}
\label{subsec:Bonsai_KWS_eval}

\begin{table}[t]
\caption{Test accuracies for DS-CNN and Bonsai tree variants on KWS. $\hat{D}$ = projected dimension, T = depth of tree.}
\label{BTKWS}
\vskip 0.15in
\begin{center}
\begin{scriptsize}
\begin{sc}
\begin{tabular}{lccccr}
\toprule
Network & Acc. (\%) & MACs & Ops & Model size\\
\midrule
DS-CNN    & 94.4 & 2.7M & 2.7M & 22.07KB \\
Bonsai ($\hat{D}$=64, T=2) & 80.20 & 0.02M & 0.02M & 140.75KB \\
Bonsai ($\hat{D}$=64, T=4) & 82.92 & 0.04M & 0.04M & 287.75KB \\
Bonsai ($\hat{D}$=128, T=2) & 81.56 & 0.04M & 0.04M & 281.5KB \\
Bonsai ($\hat{D}$=128, T=4) & 84.38 & 0.07M & 0.07M & 575.5KB \\
\bottomrule
\end{tabular}
\end{sc}
\end{scriptsize}
\end{center}
\vskip -0.1in
\end{table}

When applied to the KWS application, Bonsai shows poor prediction accuracy even
with a significantly large tree with many internal and leaf nodes. As shown in Table~\ref{BTKWS}, Bonsai trees achieve poor accuracies,
saturating at about $84\%$, even when the tree architecture is scaled up with
wider projection layers, more tree nodes and trained for longer\footnote{Bonsai trees in Table~\ref{BTKWS} are trained significantly longer than the other networks in this work. The learning rate is initially chosen as $0.001$, and later gradually reduced after every $100$ epochs.}.
Furthermore, a major fraction of the model
size (e.g. $69.63\%$ of Bonsai tree with $\hat{D}$=$64$ and T=$2$) is attributed to
the fully-connected (FC) layer used in projecting the incoming input data to
low-dimensional space.  Clearly, weight quantization\footnote{Each Bonsai tree weight in Table~\ref{BTKWS}
requires $4$ bytes to store.} and aggressive pruning will
reduce the model size further, as described in~\cite{bonsaitree2017},
however it will not be able to recover the significant accuracy loss of Bonsai
trees when compared to that of the neural models for KWS (e.g. DS-CNN). It is worth emphasizing that
although Bonsai occupies a large memory footprint, its computational costs are
very low in comparison to those neural models.

\subsubsection{The limitations of Bonsai trees for KWS}

While~\cite{bonsaitree2017}  shows the effectiveness of Bonsai trees for the
applications they considered, our results show that for more complex
applications, there might be a fundamental limitation in the expressiveness of
Bonsai trees.  More specifically, the simple projection matrix that is made of
a FC layer in a Bonsai tree is likely not effective in compressing KWS's
initial speech inputs to extract rich useful features.  This observation is
further corroborated by prior
works~\cite{helloEdge2017,CRNN_KWS2017,CNN_KWS2015} on designing
state-of-the-art neural networks for small-footprint KWS that leverages
convolutional layers instead to compress complex speech inputs of KWS
applications to extract a few rich, meaningful features. 

Based on these results, we can conclude that StrassenNets and Bonsai trees,
while effective at reducing model complexity for some models, have
limitations when applied to a representative IoT application. This
motivates the use of a potential hybrid model - one that can use the
feature extraction capabilities of a convolutional network while also reducing the amount of compute required for subsequent
classification of these features. This hybrid model proposed in this paper exploits Bonsai tree's strength as a compute-efficient classifier given rich features and couples that with StrassenNets to achieve significant reduction in model size.



\section{Hybrid Neural-Tree Architecture}
\label{sec:hybridNetwork}

We propose a hybrid neural-tree architecture that can leverage a few
convolutional layers to extract the minimal set of necessary local features,
and then can rely on powerful branching functions and non-linear Bonsai tree
nodes to find global correlation between features and to perform the required
classification. As the tree section of the hybrid network is comparatively 
compute-efficient in terms of MAC operations, use of it to find the global
interaction between local features and classifying the voice commands should
result in an overall reduction of computational costs compared to a
neural-only state-of-the-art network for KWS without compromising its accuracy. 
DS convolutional layers are used in particular for feature extraction in the
hybrid network. Additionally, the matrix multiplications associated with the entire hybrid
network are strassenified to reduce multiplications and the overall memory
footprint to enable a more compact network.

\begin{figure}[t]
   \centering
   \includegraphics[scale=0.5,clip=true,trim=0cm 0cm 0cm 0cm]{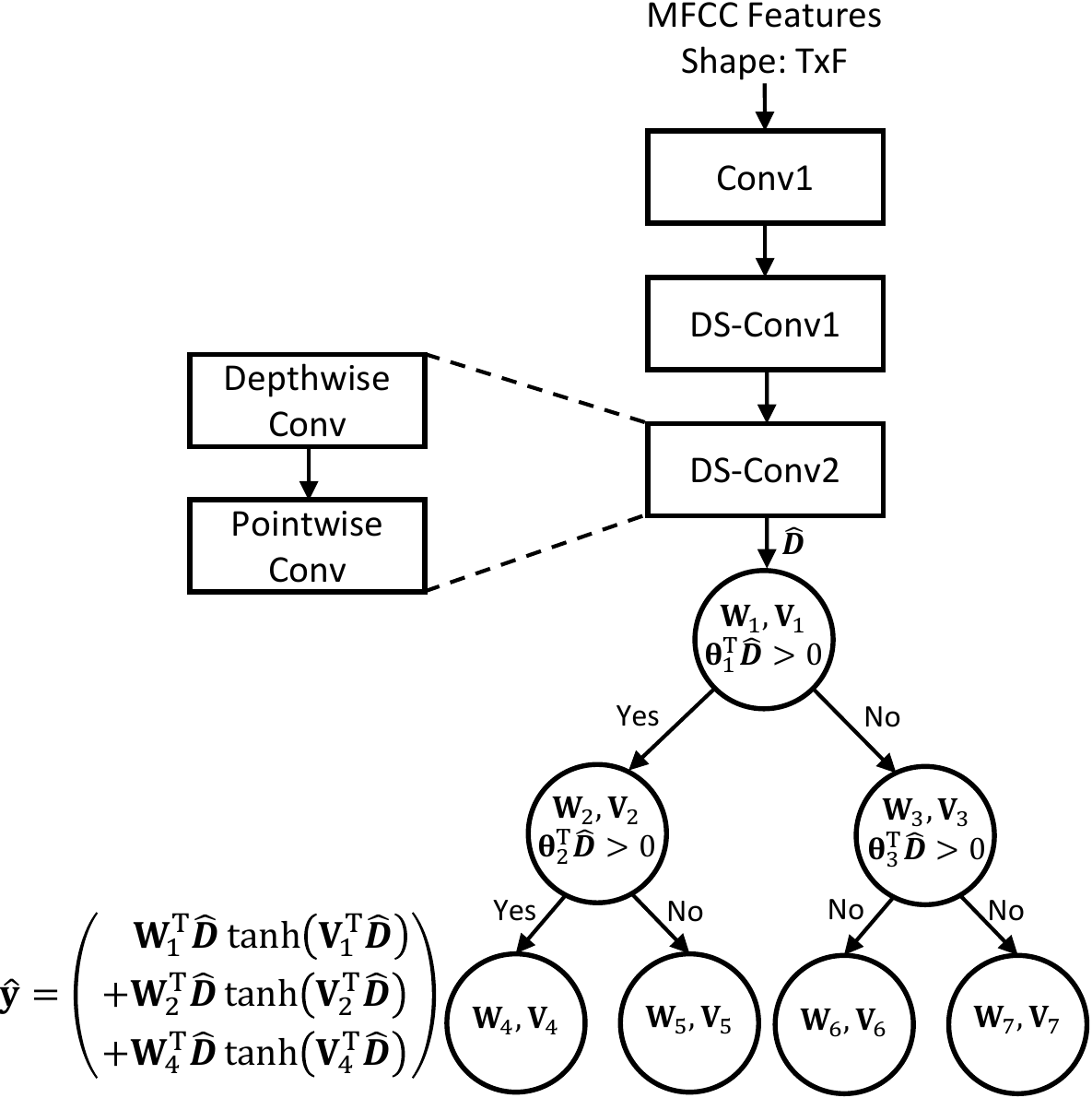}
   \caption{Hybrid neural-tree architecture.}
   \label{fig:hybrid_network}
\end{figure}

\textbf{Architecture. } 
Figure~\ref{fig:hybrid_network} shows the hybrid neural-tree architecture
optimized for KWS application with the corresponding parameters. The raw
time-domain speech signal is converted to $2$-D MFCC (Mel-frequency cepstral
coefficients) inputs for succinct representation and efficient training. Speech
features are first extracted from the MFCC inputs by one standard convolutional
layer followed by two DS convolutional layers which greatly reduce
dimensionality of the original speech signal.
The low-dimensional compressed speech features are then fed to a single depth $2$ Bonsai tree with $3$ internal and $4$ leaf nodes to provide global
interaction and to identify the appropriate keyword in the detected voice
command.

Note that the branching functions of the tree's internal nodes output a
probability to influence whether the low-dimensional speech sample should be
branched to a node's left or right child. During inference, non-linear
prediction scores are computed for all tree nodes regardless of the most
probable tree path traversed by the compressed sample. The tree nodes from the
least probable paths contribute insignificantly to the overall prediction
score. Computing prediction scores for all nodes certainly increases prediction
costs. As the hybrid network for KWS has a shallow depth $2$ tree, the
incremental costs from computing prediction scores for all nodes is marginal in
comparison to the computational costs of the overall hybrid network. 
However,
evaluating the entire tree ensures that the tree computation does not incur any
control-flow overhead in the processor from unpredictable
branching
in internal nodes. This, in turn, results in a more resource-efficient,
data-parallel computation pattern and a more efficient utilization of any available SIMD units.
For example, in Figure~\ref{fig:hybrid_network}, a low-dimensional speech feature sample
$\hat{D}$ finds the leftmost tree path most \textit{probable} to traverse and
as a result the three tree nodes along the leftmost path in the tree contribute the most to
the overall prediction score, even though the prediction scores for all tree
nodes are computed during inference.

\textbf{End-to-end training. } The three convolutional layers along with all the
tree nodes of the hybrid network are trained jointly so as to maximize accuracy.
A gradient-descent (GD) based training algorithm is used to train the hybrid
network end-to-end. Note that the path traversed by a training point in a
standard decision tree is a sharply discontinuous function of parameters of
internal branching nodes thereby making gradient based techniques ineffective.
In order to make effective use of GD algorithm with a differentiable loss
function, the training of a Bonsai tree begins with smooth activation functions
of internal branching nodes to allow points to traverse multiple paths in the
tree. As training progresses, the activation functions of internal branching nodes
are tuned to ensure that points gradually start traversing at most a single
path.

\textbf{Strassenified hybrid network. } Finally, in order to reduce the hybrid
network's memory footprint, we apply strassenified matrix multiplications to
its convolution layers and tree section to create the strassenified hybrid
network. Note that each of the tree nodes of the hybrid network learns two
matrices $W$ and $V$ to compute a non-linear prediction score. Computation of
this prediction scores at the tree nodes involves matrix multiplications, which
are strassenified as well. The hidden layer width ($r$) of the strassenified
hybrid network is set to $0.75c_{out}$ for convolution layers($c_{out}$ is the
number of output channels of a convolution layer), whereas $r$ for the tree
nodes is set to the number of targets ($L$) of the multi-class KWS
classification problem.

Training a network with strassenified matrix multiplications essentially
involves learning ternary $W_a$, $W_b$, and $W_c$ matrices for each
strassenified network layer. We employ the training procedure described in the
StrassenNets~\cite{strassennets2018} work to train the strassenified hybrid
network.  Training of the strassenified hybrid network begins with full
precision $W_a$, $W_b$, and $W_c$.  Once the network is sufficiently trained
with full-precision $W_a$, $W_b$, and $W_c$, the elements of these three
strassen matrices are quantized to ensure ternary-valued weights in them and
the training continues. Quantization converts the full-precision $W_b$ to a
ternary-valued $W_b^t$ along with a scaling factor ($W_b$ = scaling factor *
$W_b^t$). $W_a$ and $W_c$ are quantized the same way. Once the training
recovers the accuracy loss with quantized strassen matrices, the three strassen
matrices are fixed and the scaling factors associated with them are absorbed by
full-precision $vec(A)$ portion of strassenified matrix multiplication. Note that in the context of strassenified matrix multiplications of a
network layer, $A$ is associated with the weights or filters of the layer and
$B$ is associated with the corresponding activations or feature maps.  As a
result, after training, $W_a$ and $vec(A)$ can be collapsed into a vector
$\hat{a} = W_avec(A)$, as they are both fixed during inference.  We follow the
weight quantization procedure described in the
StrassenNets~\cite{strassennets2018} work for quantizing all strassen matrices
of the hybrid network.

Furthermore, in order to recover any accuracy loss of the hybrid network
compressed with strassenified matrix computations, knowledge distillation (KD)
is exploited during training, as described in~\cite{strassennets2018}\footnote{We apply KD while training the strassenified DS-CNN networks in Section~\ref{sec:modelcompression} as well. The results reported for strassenified networks in Table~\ref{strassenKWS} are obtained using KD.}. Using
KD, an uncompressed teacher network can transfer its prediction ability to a
compressed student network by navigating its training. We use the uncompressed
hybrid network as the teacher network and the compressed strassenified network
as the student network in this work.

In short, the strassenified hybrid neural-tree architecture essentially
combines the strengths of DS convolutions, Bonsai trees, and strassenified
matrix computations, with additional strategies applied during training to
improve the overall performance, while keeping a small-footprint size.

\section{Experiments and Results}
\label{sec:evaluation}

\textbf{Datasets. } We evaluate the hybrid neural-tree architecture on the
Google speech commands dataset~\cite{SpeechCommandsDataset2018} and compare it
against the state-of-the-art DS-CNN~\cite{helloEdge2017},
BinaryCmd~\cite{binaryCmd2018} and other baseline network architectures for KWS
from literature~\cite{CRNN_KWS2017,LSTM_KWS2016,CNN_KWS2015,DNN_KWS2014}
analysed in~\cite{helloEdge2017}.  The entire dataset consists of $65$K
different samples of $1$-second long audio clips of $30$ keywords, collected
from thousands of different people. The length of each audio clip is $1$
second, which is sufficiently long to capture one keyword.  The corresponding $40$
MFCC features are obtained from a speech frame of length $40$ms with a stride
of $20$ms, yielding an input dimensionality of $49\times10$ features for $1$
second of audio.  The different network architectures are trained to classify
the incoming audio into one of the $10$ keywords - ``Yes'' ``No'' ``Up'' ``Down''
``Left'' ``Right'' ``On'' ``Off'' ``Stop'' ``Go'' along with ``silence'' (i.e. no word
spoken) and ``unknown'' word, which is the remaining $20$ keywords from the
dataset.  The dataset is split into $80\%$ for training, $10\%$ for validation
and $10\%$ for testing.  Training samples are augmented by applying background
noise and random timing jitter to provide robustness against noise and
alignment errors. We follow the input data processing procedure described
in~\cite{helloEdge2017} for training the baseline and hybrid networks presented
here.

\textbf{Hybrid network training. } We use the Adam optimization algorithm to
train the networks in the Tensorflow framework~\cite{tensorflow2016}. We use multi-class
hinge loss to train the hybrid network\footnote{We use hinge loss to train the baseline Bonsai trees in Section~\ref{subsec:Bonsai_KWS_eval}, whereas we use standard cross-entropy loss to train the strassenified DS-CNN networks in Section~\ref{subsec:ST_DS-CNN_eval}.}. The Adam optimizer with hinge loss achieves marginally better accuracy
for the hybrid network than with cross-entropy loss.  The network architectures
are trained on the full training set and evaluated based on the classification
accuracy on the test set. With a batch size of $20$, the hybrid network is
trained for $135$ epochs with initial learning rate of $0.001$ and
progressively smaller learning rates after every $45$ epochs.  The training
time for the hybrid network is restricted to $135$ epochs to match against the
epochs required in training the baseline DS-CNN network. 

\textbf{Hybrid network evaluation. } The resulting testing accuracy along with
the model size and the number of multiplications, and additions in the
matrix-multiplication operations of the hybrid network is shown in
Table~\ref{results_hybridnet} and compared against prior works.  The hybrid
network achieves an accuracy of $94.54\%$ when compared to DS-CNN's accuracy of
$94.4\%$, while reducing the number of operations by
$44.4\%$.  The reduction in operations of the hybrid network stems from its
compute-efficient tree portion.  
Note that all the baseline networks in Table~\ref{results_hybridnet}
require $1$ byte to store their weights as opposed to the $4$ bytes
required to store the weights of the uncompressed hybrid
network.
Consequently, the uncompressed hybrid network requires a larger model
size of $94.25$KB when compared to other baselines in
Table~\ref{results_hybridnet}.  Clearly, straightforward quantization of the weights
of the hybrid network to low-precision values will result in reducing its model
size. However, as discussed in this work, we instead apply StrassenNets over
the entire hybrid network to reduce
its full-precision ($4$ bytes) parameters and resultant model size.

\begin{table}[t]
   \caption{Comparison of hybrid neural-tree network (HybridNet) against DS-CNN, the current state-of-the-art for KWS application, and other baselines presented in~\cite{helloEdge2017}.} 
\label{results_hybridnet}
\vskip 0.15in
\begin{center}
\begin{tiny}
\begin{sc}
\begin{tabular}{lccccr}
\toprule
	Network & Acc. (\%) & MACs & Ops & Model size\\
\midrule
DS-CNN    & 94.4 & 2.7M & 2.7M & 22.07KB \\
CRNN & 94.0 & 1.5M & 1.5M & 73.7KB\\
GRU & 93.5 & 1.9M & 1.9M & 76.3KB\\
LSTM & 92.9 & 1.95M & 1.95M & 76.8KB\\
Basic LSTM & 92.0 & 2.95M & 2.95M & 60.9KB\\
CNN & 91.6 & 2.5M & 2.5M & 67.6KB\\
DNN & 84.6 & 0.08M & 0.08M & 77.8KB\\
HybridNet & 94.54 & 1.5M & 1.5M & 94.25KB \\
\bottomrule
\end{tabular}
\end{sc}
\end{tiny}
\end{center}
\vskip -0.1in
\end{table}

\textbf{Strassenified hybrid network training. } We begin by training the
strassenified hybrid network (ST-HybridNet) with full-precision strassen matrices ($W_a$,
$W_b$, and $W_c$) for $135$ epochs.  The learning rate is initially chosen as
$0.001$, and later gradually reduced after every $45$ epochs.  We then activate
quantization for these strassen matrices and the training continues.
Finally, we fix the strassen matrices
to their learned
ternary values and continue training for another $135$ epochs to ensure that
the scaling factors associated with these matrices can be absorbed by
full-precision $vec(A)$ portion of strassenified matrix multiplication.

\textbf{Strassenified hybrid network evaluation. } The testing accuracy of the
ST-HybridNet is shown in
Table~\ref{results_strassen_hybridnet}, along with the reduction in
the number of operations, and the model size in comparison to the
uncompressed hybrid network.  The ST-HybridNet achieves similar
accuracy to that of the uncompressed hybrid network and the baseline DS-CNN
while reducing the number of multiplications and additions by $98.89\%$ and
$12.22\%$, respectively, over the baseline DS-CNN network.  Of particular note is that it
reduces the number of additions to about $2.37$M when compared to $4.09$M
additions of strassenified DS-CNN network described in
Section~\ref{sec:modelcompression}. This, in turn, results in fewer
overall operations, $2.4$M, for the ST-HybridNet when compared to $2.7$M operations
of the baseline DS-CNN and $4.15$M operations of the strassenified DS-CNN. 
This reduction in operations is primarily attributed to strassenifying a few (three) convolutional layers and a compute-efficient tree as opposed to strassenifying all of the five convolutional layers found in the baseline DS-CNN model. 
Owing to the ternary weights matrices,
the ST-HybridNet reduces the model size to $14.99$KB when
compared to $22.07$KB of the baseline DS-CNN network thus enabling a $32.1\%$ savings in model size for KWS\footnote{During inference, the batch normalization parameters (beta, moving mean, and moving variance) are folded either into the full-precision bias parameters of the preceding convolution layers and/or into the full-precision $vec(A)$ parameters of the ST-HybridNet.}.
Furthermore, our ST-HybridNet does not incur any accuracy loss over the baseline DS-CNN
while achieving reduction in computational costs and model size.
The use of KD in training the ST-HybridNet does not result in any tangible change in accuracy. 

\begin{table}[t]
   \caption{Comparison of the strassenified hybrid neural-tree network (ST-HybridNet) against the uncompressed hybrid network, DS-CNN, and strassenified DS-CNN network (ST-DS-CNN) presented in Section~\ref{sec:modelcompression}.}
\label{results_strassen_hybridnet}
\vskip 0.15in
\begin{center}
\begin{tiny}
\begin{sc}
\begin{tabular}{lccccccr}
\toprule
	Network & Acc. & Muls, Adds & MACs & Ops & Model\\
	&  (\%) &  &  &  & size\\
\midrule
DS-CNN    & 94.4 & - & 2.7M & 2.7M & 22.07KB \\
ST-DS-CNN & 94.09 & 0.06M, 4.09M & - & 4.15M & 19.26KB\\
($r = 0.75c_{out}$) &  &  &  &  & \\
HybridNet & 94.54 & - & 1.5M & 1.5M & 94.25KB\\
ST-HybridNet & 94.51 & 0.03M, 2.37M & - & 2.4M & 14.99KB \\
(without KD) &  &  &  &  & \\
ST-HybridNet & 94.41 & 0.03M, 2.37M & - & 2.4M & 14.99KB\\
(with KD) &  &  &  &  & \\
\bottomrule
\end{tabular}
\end{sc}
\end{tiny}
\end{center}
\vskip -0.1in
\end{table}

We perform exhaustive search of feature extraction hyperparameters and model hyperparameters to develop ST-HybridNet. Table~\ref{hybridnet_hyperparams_search} summarizes the  hyperparameters of different configurations of ST-HybridNet along with their impact on accuracy and computational complexity. 
A ST-HybridNet with two convolutional layers (one standard convolutional layer followed by one DS convolutional layer) and  a single depth
$2$ Bonsai tree with $7$ nodes ($3$ internal and $4$ leaf nodes) reduces computational requirements of a KWS model 
but at the cost of more than $3\%$ accuracy loss in comparison to the baseline DS-CNN network. Even a ST-HybridNet with three convolutional layers and a depth
$1$ Bonsai tree with $3$ nodes ($1$ internal and $2$ leaf nodes) cannot preserve the baseline accuracy. This hyperparameter search subsequently results in designing the ST-HybridNet with three convolutional layers and a depth $2$ Bonsai tree for KWS application in this work.

\begin{table}[t]
   \caption{Different network hyperparameters of ST-HybridNet and their impact on accuracy and number of operations. D = depth of tree, N = number of tree nodes.}
\label{hybridnet_hyperparams_search}
\vskip 0.15in
\begin{center}
\begin{scriptsize}
\begin{sc}
\begin{tabular}{lcccr}
\toprule
	Network & Model & Acc. (\%) & Ops\\
	& Hyperparameters & & \\
\midrule
ST-HybridNet & 2 convolutional & 91.1 & 1.53M\\
& layers, D=2, N=7 & & \\
ST-HybridNet & 3 convolutional & 93.15 & 2.39M\\
& layers, D=1, N=3 & & \\
ST-HybridNet & 3 convolutional & 94.51 & 2.4M\\
& layers, D=2, N=7 & & \\
\bottomrule
\end{tabular}
\end{sc}
\end{scriptsize}
\end{center}
\vskip -0.1in
\end{table}

As ternary $W_a$ and full-precision $vec(A)$ weights of the ST-HybridNet are both fixed during inference, they are learned jointly as collapsed full-precision $\hat{a}$ ($W_avec(A)$) from scratch, and these full-precision $\hat{a}$ weights along with the bias parameters occupy $7.34$KB out of $14.99$KB of the ST-HybridNet.
The limited memory of microcontroller systems motivates further reducing the numerical precision of the entire network model, including the inputs, outputs, activations, and remaining full-precision weights, to minimize the overall memory footprint during inference.

\textbf{Quantization of activations and remaining full-precision weights of strassenified hybrid network. }
Quantization can convert these high-precision floating-point weights and activations of ST-HybridNet to a low-precision fixed-point format more amenable for deployment in resource-constrained microcontrollers. This can also ensure faster inference through the use of fixed-point integer operations rather than floating-point operations.

We follow the quantization procedure described in~\cite{Qiu2016,helloEdge2017} for quantizing the remaining full-precision weights and activations of the pre-trained ST-HybridNet. The full-precision weights and activations of the pre-trained ST-HybridNet are quantized progressively, one layer at a time, by finding the optimal min/max range for each layer that minimizes the loss in accuracy because of quantization. Table~\ref{results_quantized_strassen_hybridnet} captures the accuracy, model size and total memory required for storing the weights and activations of the quantized ST-HybridNet model during inference.
We assume that the memory for activations is reused across different layers and, hence, the memory requirement for the activations uses the maximum of two consecutive layers (output activations from a preceding layer and input activations to the following layer).
As shown in Table~\ref{results_quantized_strassen_hybridnet}, quantizing activations of our ST-HybridNet to $8$ bits reduces the model size to $10.54$KB and the total memory footprint to $26.17$KB, albeit with a very small loss in accuracy of $0.27\%$.
This is primarily attributed to the intermediate activations (activations produced post-convolution with strassen matrix $W_b$) of the strassenified depthwise convolutions of two DS layers that require $16$ bits to represent their range precisely and preserve baseline accuracy. 
Quantizing the $\hat{a}$ weights and the intermediate activations of the depthwise convolution layers to $16$ bits and the remaining full-precision weights and activations to $8$ bits in our ST-HybridNet 
not only recovers the small accuracy loss of the quantized ST-HybridNet with fully $8$ bits activations, but also achieves marginally better accuracy than the baseline quantized DS-CNN network, possibly owing to better regularization because of quantization.
The quantized ST-HybridNet with mixed $8/16$ bits activations ($16$ bits for strassenified depthwise convolution layers) reduces model size to $10.54$KB and requires an overall memory footprint of $41.8$KB. Out of $41.8$KB of total footprint, $31.25$KB of memory is primarily attributed to the storage of $16$ bits intermediate activations of strassenified depthwise layers. Remaining layers of ST-HybridNet require at most $15.63$KB of memory during inference for storing activations of two consecutive layers.  

{\em In summary, the quantized ST-HybridNet reduces model size by $52.2\%$ and overall memory footprint by $30.6\%$ while incurring a negligible loss in accuracy when compared to the baseline quantized DS-CNN.}
It is important to note that Table~\ref{results_quantized_strassen_hybridnet} captures accuracy results with quantizing weights and activations of the \textit{pre-trained} ST-HybridNet. In other words, the ST-HybridNet here is not retrained post quantization. We believe this $0.27\%$ drop in accuracy with quantizing activations to $8$ bits can be recovered via integrating the quantization process into the training procedure of ST-HybridNet. We leave this exploration for future work.

\begin{table}[t]
   \caption{Model quality and memory footprint after quantizing weights and activations of pre-trained ST-HybridNet. Memory footprint denotes the total memory required for storing weights and activations of a network during inference. 1KB = 1024 bytes.}
\label{results_quantized_strassen_hybridnet}
\vskip 0.15in
\begin{center}
\begin{tiny}
\begin{sc}
\begin{tabular}{lcccccr}
\toprule
	Network & Acc. & Ops & Model & Total memory \\
	&  (\%) &  &  size & footprint\\
\midrule
DS-CNN    & 94.4 & 2.7M & 22.07KB & 37.7KB \\
ST-HybridNet Quantized & 94.13 & 2.4M & 10.54KB & 26.17KB\\
(fully 8b activations) &  &  &  & \\
ST-HybridNet Quantized & 94.71 & 2.4M & 10.54KB & 41.8KB\\
(mixed 8b/16b activations) &  &  &  & \\
\bottomrule
\end{tabular}
\end{sc}
\end{tiny}
\end{center}
\vskip -0.1in
\end{table}

\section{Comparative Analysis}
\label{sec:related_work}
In recent years, numerous research efforts have been devoted to compressing neural networks for deployment in resource-constrained environments through the use of model pruning, quantization, low-rank matrix factorization, compact network architecture design, etc. ST-HybridNet falls into the category of compact architecture design for IoT applications. In order to demonstrate the efficacy of ST-HybridNet over other model compression techniques, we apply state-of-the-art pruning and quantization techniques to the baseline DS-CNN network and present their performance in this section.

\textbf{Model pruning. }
Pruning away unimportant connections induces sparsity in a neural network, thereby reducing the number of nonzero-valued parameters in the model. 
Recent works ~\cite{DeepCompression2016,Narang2017a,Zhu2017} on model pruning have shown that common networks
have significant redundancy and can be pruned dramatically during training with marginal to no degradation in the model accuracy.
By reducing nonzero parameters of a network, model pruning attempts to reap improvements in inference time and energy-efficiency.
In addition to the storage for the nonzero model elements, a pruned model requires to store auxiliary data structures for indexing these elements resulting in additional storage overhead. 
On top of that the specialized routines involved with sparse matrix computations of a pruned model require considerable sparsity in associated matrices to realize any benefit in runtime owing to their irregular computation pattern and under utilization of any available SIMD units. Typically, a sparsity level of $70\%$ or above is required in order for a sparse matrix computation to observe any benefit in runtime than the corresponding dense matrix computation.




We follow the gradual pruning technique proposed in~\cite{Zhu2017} to prune the parameters of the baseline DS-CNN network. \cite{Zhu2017} gradually prunes the small magnitude weights to achieve a preset level of network sparsity. Table~\ref{pruning_dscnn} compares the performance of sparse DS-CNN models pruned to varying extents.
As shown in Table~\ref{pruning_dscnn}, although a $50\%$ sparse DS-CNN model causes marginal loss in accuracy, it will be hard for the DS-CNN model to realize any benefit with this sparsity either in runtime, due to the sparse matrix computation, or in model size, due to the overhead from storing indices when compared to ST-HybridNet. Nevertheless, as different model pruning techniques~\cite{Guo2016,Aghasi2017,Wen2016,He2017,Luo2017,Yang2018,Gordon2018} are orthogonal to our compression scheme, they can be used in conjunction with ST-HybridNet to further reduce model size.

\begin{table}[t]
   \caption{Model size and accuracy tradeoff for  DS-CNN, the current state-of-the-art for KWS application.}
\label{pruning_dscnn}
\vskip 0.15in
\begin{center}
\begin{scriptsize}
\begin{sc}
\begin{tabular}{lcccr}
\toprule
	Sparsity & Nonzero parameters & Acc. (\%)\\
\midrule
0\% & 23.18K & 94.4 \\
50\% & 11.59K & 94.03 \\
75\% & 5.79K & 92.37 \\
90\% & 2.31K & 87.41 \\
\bottomrule
\end{tabular}
\end{sc}
\end{scriptsize}
\end{center}
\vskip -0.1in
\end{table}

\textbf{Model quantization. }
As mentioned previously, the baseline DS-CNN network in Table~\ref{results_hybridnet} uses an $8$-bit fixed-point quantized format to represent weights. In order to observe the impact of binary/ternary quantization~\cite{Courbariaux2015,XNORNet2016,ABCNet2017,Cai2017,ternaryWeightNetworks2016,TTQ2016}, we apply ternary weight quantization~\cite{ternaryWeightNetworks2016} over the baseline DS-CNN network. Ternary quantization of the weights of DS-CNN reduces the model size to $9.92$KB but drops prediction accuracy significantly (by $2.27\%$). Any increase in the size of the DS-CNN network to recover the accuracy loss while using ternary quantization will lead to an increase in the number of MAC operations.
Recent work on BinaryCmd~\cite{binaryCmd2018} achieves significant reduction in KWS model size but at the cost of $3.4\%$ accuracy loss compared to the baseline DS-CNN network. 

\textbf{Low-rank matrix factorization. }
Besides pruning and quantization, low-rank matrix factorization techniques
~\cite{Jaderberg2014,Tai2015,Wen2017} exploit parameter redundancy to obtain low-rank approximations of weight matrices without compromising model accuracy. Strassen matrices of our ST-HybridNet can adopt these prior proposals to further reduce model size and computational complexity.

\textbf{Compact network architectures. }
Much research has been done in recent years on developing compact architectures~\cite{DNN_KWS2014,CNN_KWS2015,LSTM_KWS2016,CRNN_KWS2017,helloEdge2017,BoLi2017,binaryCmd2018,Myer2018,Coucke2018} for keyword spotting on resource-constrained environments.
Recent work on EdgeSpeechNets~\cite{EdgeSpeechNets2018} produces good results albeit with significantly higher computational complexity.
It is targeted for mobile processors (Arm Cortex-A53) as it requires at least $10x$ more MAC operations than our baselines and proposed ST-HybridNet, all of which are are primarily targeted for microcontrollers. EdgeSpeechNet is well beyond the constrained compute and storage budget of typical microcontrollers described in~\cite{helloEdge2017}.

\section{Conclusion and Future Work}
\label{sec:conclusion}

We have presented a hybrid network architecture for a keyword spotting
application capable of giving start-of-the-art accuracy levels while requiring
a fraction of the model
parameters and considerably fewer operations per inference pass. 
The hybrid architecture makes this possible by leveraging a few neural
DS layers to extract features from the audio input and feeding
those features to a
shallow Bonsai decision tree to perform the classification.
Furthermore, StrassenNets is used to significantly reduce the model size. The
reduction in computation from the Bonsai tree, the parameter-efficiency of the
DS convolutional layers, and the model footprint reduction
provided by StrassenNets all combine to make the KWS model much more amenable
to run on a highly constrained IoT device.

In the next iterations of this work, we will explore different algorithmic ways
to constrain the number of additions in a strassenified network dominated with
DS layers or specifically pointwise convolutions (e.g. MobileNets architecture)
and develop architectures or specialized hardware suitable for such changes.
This will not only enable a more homogeneous network architecture, but also
will pave the way for incorporating StrassenNets into the next generation
microcontrollers while maintaining acceptable computational costs and model
size. We leave this exploration for future work.





\bibliography{example_paper}
\bibliographystyle{sysml2019}



\end{document}